\documentclass[journal,twoside,web]{ieeecolor}
\usepackage{tmi}
\usepackage{cite}
\usepackage{amsmath,amssymb,amsfonts}
\usepackage{algorithm,algorithmic}
\usepackage{graphicx}
\usepackage{textcomp}

\usepackage{titlesec}
\usepackage{multirow}
\usepackage{setspace}
\usepackage{array}
\usepackage{soul}
\usepackage{threeparttable}
\usepackage{makecell}
\usepackage{hyperref}  
\def\BibTeX{{\rm B\kern-.05em{\sc i\kern-.025em b}\kern-.08em
    T\kern-.1667em\lower.7ex\hbox{E}\kern-.125emX}}
\begin{document}
\definecolor{mygray}{gray}{.93}
	\definecolor{mycolor_light_blue}{RGB}{111, 185, 211}
	\definecolor{lightblue}{RGB}{0, 153, 255}
\title{ScribbleVS: Scribble-Supervised Medical Image Segmentation via Dynamic Competitive Pseudo Label Selection}
\author{Tao Wang, Xinlin Zhang, Zhenxuan Zhang, Yuanbo Zhou, Yuanbin Chen, Longxuan Zhao, Chaohui Xu, Shun Chen, Guang Yang, \IEEEmembership{Senior Member, IEEE}, and Tong Tong
\thanks{This work was supported in part by National Natural Science Foundation of China under Grant 62171133; in part by the ERC IMI under grant 101005122, the H2020 under Grant 952172; in part by MRC under Grant MC/PC/21013; in part by the Royal Society under Grant IEC\textbackslash NSFC\textbackslash 211235, in part by the NVIDIA Academic Hardware Grant Program, in part by the SABER project supported by Boehringer Ingelheim Ltd, in part by the NIHR Imperial Biomedical Research Centre under Grant RDA01; in part by the Wellcome Leap Dynamic Resilience, in part by the UKRI guarantee funding for Horizon Europe MSCA Postdoctoral Fellowships under Grant EP/Z002206/1; in part by the UKRI MRC Research Grant, TFS Research Grant MR/U506710/1; in part by the UKRI Future Leaders Fellowship under Grant MR/V023799/1; in part by the Guiding Projects of Fujian Provincial Technology Research and Development Program under Grant2022Y0023, and in part by Joint Funds for the innovation of science and Technology, Fujian province under Grant 2024Y9264.
 }
\thanks{Tao Wang is with the college of physics and information engineering, University of Fuzhou University, Fuzhou 350108, China; Department of Bioengineering and Imperial-X, Imperial College London, United Kingdom (221110031@fzu.edu.cn).} 
\thanks{Xinlin Zhang, Yuanbo Zhou, Yuanbin Chen, and Tong Tong are with the college of physics and information engineering, University of Fuzhou University, Fuzhou 350108, China (xinlin@fzu.edu.cn; 221110034@fzu.edu.cn; 241110034@fzu.edu.cn; ttraveltong@fzu.edu.cn).}
\thanks{Chaohui Xu is with the Fuzhou University Zhicheng College, Fuzhou 350108, China(xch0033@163.com).} 
\thanks{Longxuan Zhao is with the Department of Medical Bioinformatics, School of Basic Medical Sciences, Peking University Health Science Center, Beijing, China (zhaoxuanlong254@gmail.com).}
\thanks{Zhenxuan Zhang is with the Fuzhou University Zhicheng College, Imperial College London, United Kingdom (zz2524@ic.ac.uk).} 
\thanks{Shun Chen is with the Department of Ultrasound, Fujian Medical University Union Hospital, Fuzhou, China; Fujian Medical Ultrasound Research Institute, Fuzhou, China.} 
\thanks{Guang Yang is with the Department of Bioengineering and Imperial-X, Imperial College London, United Kingdom; National Heart and Lung Institute, Imperial College London, United Kingdom; Cardiovascular Research Centre, Royal Brompton Hospital, United Kingdom; School of Biomedical Engineering and Imaging Sciences, King’s College London, United Kingdom (g.yang@imperial.ac.uk).} 
\thanks{Corresponding author: Tong Tong (email: ttraveltong@fzu.edu.cn), Shun Chen (email: shunzifjmu@fjmu.edu.cn)}}

\maketitle

\begin{abstract}
	In clinical medicine, precise image segmentation can provide substantial support to clinicians. However, obtaining high-quality segmentation typically demands extensive pixel-level annotations, which are labor-intensive and expensive. Scribble annotations offer a more cost-effective alternative by improving labeling efficiency. Nonetheless, using such sparse supervision for training reliable medical image segmentation models remains a significant challenge. Some studies employ pseudo-labeling to enhance supervision, but these methods are susceptible to noise interference. To address these challenges, we introduce ScribbleVS, a framework designed to learn from scribble annotations. We introduce a Regional Pseudo Labels Diffusion Module to expand the scope of supervision and reduce the impact of noise present in pseudo labels. Additionally, we introduce a Dynamic Competitive Selection module for enhanced refinement in selecting pseudo labels. Experiments conducted on the ACDC, MSCMRseg, WORD, and BraTS2020 datasets demonstrate promising results, achieving segmentation precision comparable to fully supervised models. The codes of this study are available at {https://github.com/ortonwang/ScribbleVS}.
\end{abstract}

\begin{IEEEkeywords}
Medical image segmentation, Scribble annotation, Pseudo labels, Weakly-supervised learning 
\end{IEEEkeywords}

\section{Introduction}
\label{sec:introduction}
\IEEEPARstart{D}{eep} neural networks have demonstrated superior effectiveness in medical image segmentation \cite{yuan2023effective}, but their dependence on densely annotated datasets remains a major barrier to clinical use. Creating pixel-level masks is labor-intensive, demanding careful boundary tracing by specialists and limiting scalability \cite{tang2024semi}. In contrast, scribble annotations can be produced in seconds with simple strokes that better align with clinical routines. These annotations ease the burden on expert radiologists and can be generated accurately by junior clinicians, substantially reducing labeling costs while providing sufficient information to train reliable models.

\begin{figure}[!t]
	\centering
	\includegraphics[width=0.45\textwidth]{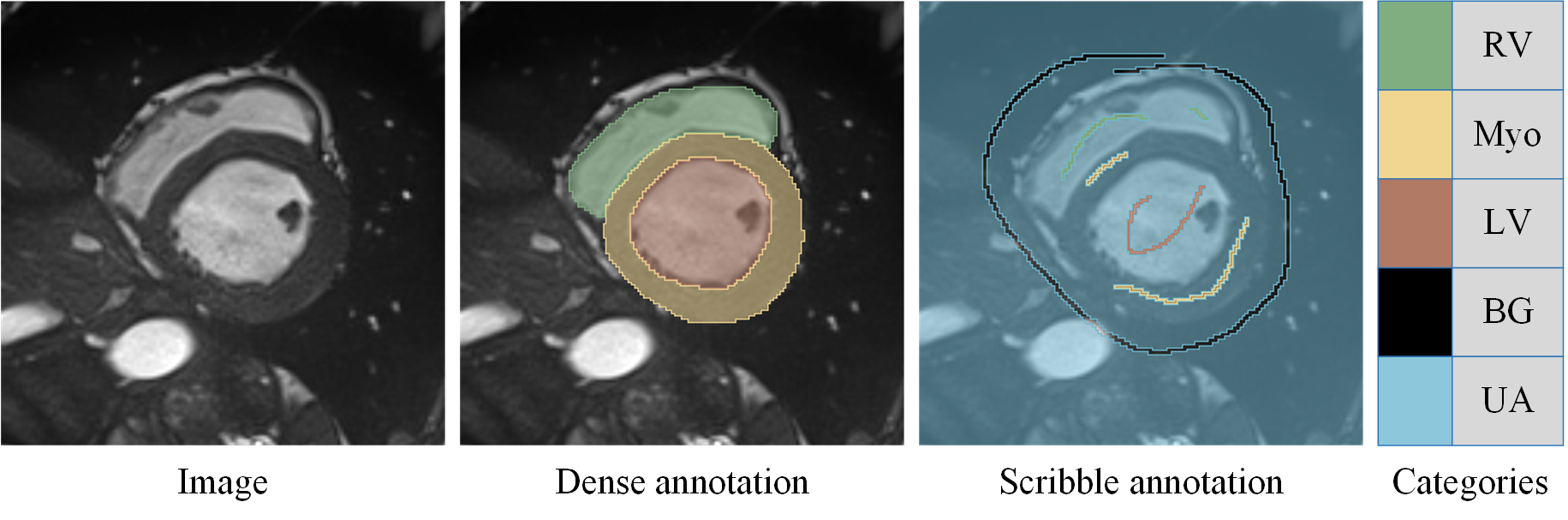}		
	\caption{Examples of pixel level annotation and scribble annotations. BG, RV, Myo, LV, and UA represent the background, right ventricle, myocardium, left ventricle, and unannotated pixels respectively.}
	\label{figure0}
\end{figure}

Despite these advances, the success of such methods still heavily relies on large-scale, finely detailed manual annotations, which are labor-intensive. In medical imaging, annotating a single image can take hours, requiring substantial domain expertise and time \cite{DMPLS}. To alleviate the annotation burden, some approaches have explored semi-supervised learning (SSL), which incorporates unlabeled data into model training \cite{luo2022semi}.
However, SSL still requires a subset of fully labeled images, which continues to impose significant annotation costs. In response, researchers have turned to weakly supervised segmentation using sparse annotations (e.g., scribbles \cite{Lin2016CVPR}, bounding boxes \cite{tian2021boxinst}, point-level annotations \cite{cheng2022pointly}) to improve labeling efficiency \cite{TAJBAKHSH2020101693}. While image-level supervision has been investigated for segmentation \cite{9440699}, such methods typically depend on large training datasets and tend to underperform when applied to limited medical data. In contrast, scribbles annotations are suitable for annotating nested structures and are easier to acquire in practice. As shown in Fig. \ref{figure0}, they offer significantly greater efficiency than dense manual annotations.

Prior research has highlighted the potential of scribble annotations in both semantic and medical image segmentation tasks \cite{DMPLS}. Motivated by this, we explore a weakly supervised segmentation approach that exclusively utilizes scribble annotations, presenting a more cost-effective solution for medical image labeling. Several studies have explored the use of scribble annotations in various application domains.
Tang \textit{et al.} \cite{Tang_2018_ECCV} introduced Conditional Random Field (CRF) regularization loss into the training of segmentation networks. In medical imaging, Can \textit{et al.} \cite{LearningtoSegment} proposed an iterative framework for model training based on scribble annotations. 
Lee \textit{et al.} \cite{Scribble2Label} employed a combination of pseudo labeling and label filtering strategy to improve label quality during training. Liu \textit{et al.} \cite{liu2022weakly} developed a unified weakly supervised framework for training networks from scribble annotations, which consists of an uncertainty-aware mean teacher and a transformation-consistency strategy. 
Valvano \textit{et al.} \cite{9389796} introduced a scribble-based segmentation model using a multi-scale Generative Adversarial Network (GAN) and attention gates mechanism, along with a novel unpaired segmentation mask strategy that requires additional annotation masks for training. Simultaneously, Luo \textit{et al.} \cite{han2024dmsps} developed a dual-branch framework, employing dynamically mixed pseudo label supervision (DMSPS) for scribble-annotated medical image segmentation. Further, Cyclemix \cite{cyclemix} was employed to generate blended images, reinforcing the training goal with consistency losses to address inconsistent segmentations. Li \textit{et al.} presented ScribbleVC \cite{ScribbleVC}, integrating scribble-based methods with segmentation networks and class embedding modules for enhanced segmentation masks.

While scribble annotations reduce the need for exhaustive expert labeling, their limited supervisory signals may compromise model accuracy. To address this, methods such as DMSPS \cite{han2024dmsps} and ScribbleVC \cite{ScribbleVC} attempt to enhance supervision through pseudo-labeling strategies. However, because scribble annotations provide only sparse and coarse supervision, the pseudo-labels generated from them often contain significant noise. Scribbles usually cover only limited regions of interest, leaving large areas of the image without explicit guidance. In these unlabeled regions, models are forced to extrapolate uncertain predictions, which frequently introduce false positives or false negatives. Moreover, the coarse nature of scribble strokes offers little boundary precision, leading to over- or under-segmentation of fine anatomical structures. When such noisy pseudo-labels are repeatedly used during training, their errors can accumulate and be reinforced, ultimately degrading the model’s segmentation accuracy and reliability.

To address this, we propose ScribbleVS, a framework specifically designed for scribble-supervised medical image segmentation. Similar to DMSPS \cite{han2024dmsps}, our method also relies on pseudo labels and is therefore not entirely immune to label noise. However, ScribbleVS introduces a strategy to refine the pseudo labels during training, thereby reducing the propagation of erroneous supervision signals. Specifically, ScribbleVS introduces the Regional Pseudo-Label Diffusion Module to reduce the impact of noisy pseudo labels and improve model robustness. Rather than treating all regions equally, the module adaptively propagates label information within local neighborhoods, guided by prediction confidence. Unreliable regions are selectively ignored, while more confident predictions influence their uncertain neighbors. This mechanism enables the model to better utilize regional context and increases its resistance to local noise within the pseudo labels. Additionally, we introduce a Dynamic Competitive Selection Module. Built upon a mean-teacher framework, this module compares the prediction accuracy of the teacher and student models within scribble-annotated regions during training. The more accurate predictions are dynamically selected and refined through the Regional Pseudo-Label Diffusion Module to generate high-quality pseudo labels, further enhancing the reliability of supervision.

The main contributions of this study are outlined below:

\begin{itemize}
	\item We introduce the ScribbleVS framework, which includes a Regional Pseudo-Label Diffusion Module designed to broaden supervision coverage. This module effectively integrates pseudo-labels while reducing the impact of noise.
	\item We design a Dynamic Competitive Selection Module that enhances model robustness by selectively using appropriate pseudo labels for training.
	\item Experiments conducted on the ACDC, MSCMRseg, WORD, and BraTS2020 datasets reveal that our approach consistently outperforms existing state-of-the-art methods, achieving segmentation precision comparable to or exceeding that of fully supervised models.
\end{itemize}
\begin{figure*}[!h]
	\centering
	\includegraphics[width=1\textwidth]{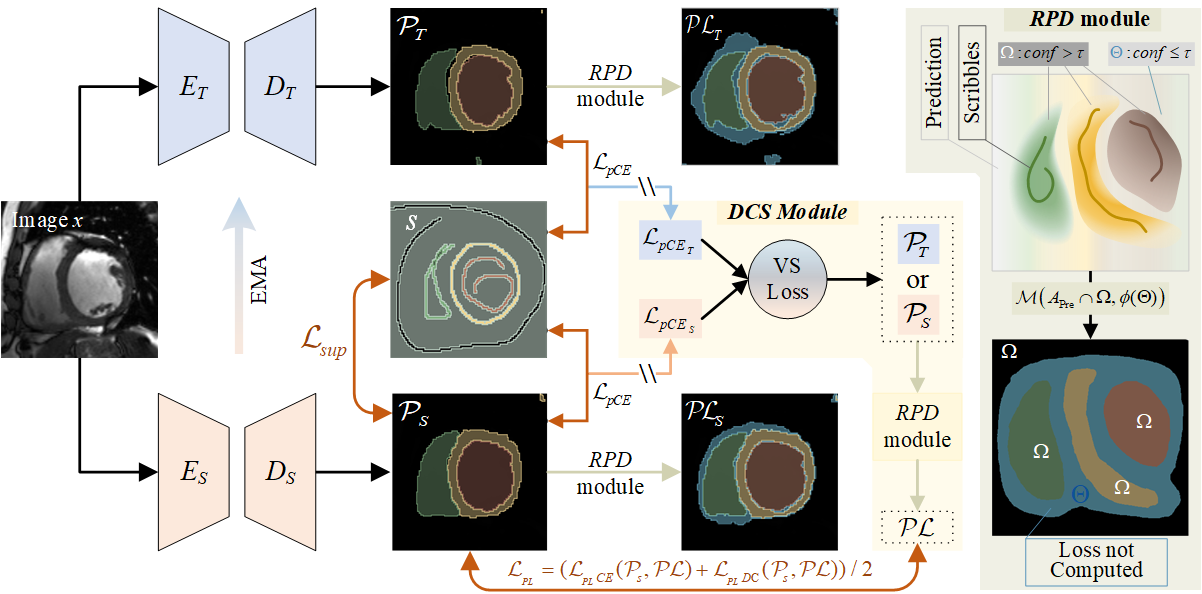}		
	\caption{The schematic diagram of the proposed ScribbleVS framework, which comprises the $RPD$ and the $DCS$ module. The $\mathcal{PL}$ denotes the pseudo labels, and the \textcolor{mycolor_light_blue}{\rule{0.2cm}{0.2cm}} areas in the $\mathcal{PL}$ indicate regions excluded from loss computation.  EMA refers to the exponential moving average mechanism, through which the teacher network is updated based on the student network’s parameters.}
	\label{figure1}
\end{figure*}

\section{Methods}
Fig. \ref{figure1} presents an overview of ScribbleVS, a framework based on a mean-teacher architecture that integrates two key modules: Regional Pseudo Label Diffusion ($RPD$) and Dynamic Competitive Selection ($DCS$). During the training process, the teacher network remains non-trainable and is updated via an exponential moving average of the student network’s parameters. The initial learning is conducted using scribble annotations through $\mathcal{L}_{sup}$. Subsequently, the $RPD$ Module generates pseudo labels to extend the supervisory signal beyond the annotated regions. The $DCS$ module further refines the learning process by dynamically selecting the more reliable pseudo labels during training. A detailed description of each module is provided in the following subsections.

\subsection{Scribble Supervision Module}
For scribble-based learning, images and their corresponding scribble annotations $s$ are utilized. As scribble annotations exclusively identify pixels with specific classes or unknown labels, the model is trained by minimizing the partial cross-entropy loss ($\mathcal{L}_{pCE}$), denoted as $\mathcal{L}_{sup}$ in Fig. \ref{figure1}. The specific formula is defined as follows:

\begin{equation}
	\mathcal{L}_{pCE}(p,s)= - \sum_{i \in \Omega_{l}} \sum_{k\in K}s[i,k] \cdot \log p[i,k]\,,
	\label{equ1}
\end{equation}
where $K$ is the label index set, and $\Omega_{l}$ denotes the set of labeled pixels in the scribble $s$. Here, $s{[i,k]}$ and $p{[i,k]}$ represent the probability that the $i$-th pixel belongs to the $k$-th category in the scribble and the prediction, respectively.
\subsection{Regional Pseudo Labels Diffusion Module}
Unlike prior methods that generate pseudo labels for all pixels indiscriminately \cite{han2024dmsps}, we introduce a Regional Pseudo Label Diffusion ($RPD$) module. This module aims to refine pseudo-labels by leveraging prediction confidence. We begin by normalizing the network prediction as follows:
\begin{equation}
	Pre(x) = \mathrm{Softmax}(f_{\theta}(x))\,,
\end{equation}
where $\theta$ denotes the parameters of the network, and $x$ represents the input image. Departing from conventional sharpening-based strategies, we partition the image domain $\mathcal{X}$ into high- and low-confidence regions based on the prediction confidence $Pre(x)$. 
The partition is defined as:
\begin{align}
	\Omega &= \{x \in \mathcal{X} \mid {Pre}(x) > \tau\}, \\
	\Theta &= \{x \in \mathcal{X} \mid {Pre}(x) \leq \tau\}, 
\end{align}
where $\mathcal{X}$ denotes the image domain, i.e., the full set of pixel locations in an image. The region $\Omega$ contains high-confidence pixels suitable for direct supervision, while $\Theta$ comprises uncertain regions excluded from direct loss computation. Subsequently, we determine the initial hard labels by the Argmax operation:
\begin{equation}
	A_{Pre} = Argmax(Pre(x))\,.
\end{equation}
To construct the final pseudo-label map, we integrate the confident predictions within $\Omega$ and the estimated pseudo labels for $\Theta$ using a regional fusion operator:
\begin{equation}
	\mathcal{PL} = \mathcal{M}\left( A_{\mathrm{Pre}} \cap \Omega, \ \phi(\Theta) \right),
\end{equation}
where $\phi$ denotes an operator that encodes the region $\Theta$ as inactive in the pseudo labels (thus ignored during loss computation), $\mathcal{M}(\cdot, \cdot)$ denotes a region-wise fusion operator that assigns labels from $A_{\mathrm{Pre}}$ in $\Omega$while preserving the inactive state of $\Theta$. This selective fusion mechanism mitigates the influence of unreliable regions, thereby reducing noise in the pseudo labels. By leveraging spatial diffusion guided by high-confidence predictions, the $RPD$ module effectively propagates supervision from sparse scribble annotations to unlabeled regions, expanding the annotation coverage from partial to dense pixel-wise guidance. The evolution of the generated pseudo labels is illustrated in Fig.~\ref{figure4}.

\begin{algorithm}[!t]
	\textbf{Require:}  \\
	\hspace*{0em}The image $x$ and corresponding scribble annotation $s$. \\
	\hspace*{0em}The Teacher Network $f_{T}(\cdot)$ and the Student Network: $f_{S}(\cdot)$.
	\hspace*{0em}1: \quad	$\mathcal{L}_{pCE_{S}} \leftarrow 0;  \hspace{6.2em}    \mathcal{L}_{pCE_{T}} \leftarrow 0$  \\
	\hspace*{0em}2: \quad $\mathcal{P}_{S}  \leftarrow f_{S}(x)$, \hspace*{5.5em} $\mathcal{P}_{T}  \leftarrow f_{T}(x))$\\
	\hspace*{0em}3: \quad $L_{pCE_{S}}  \leftarrow \mathcal{L}_{pCE}(\mathcal{P}_{S},s)$, \hspace*{1.3em} $L_{pCE_{T}}  \leftarrow \mathcal{L}_{pCE}(\mathcal{P}_{T},s)$
	\\
	\hspace*{0.0em}4: \quad \textbf{if} $L_{pCE_{S}} \leq L_{pCE_{T}}$\hspace*{3.5em} \textbf{then} 	\\
	\hspace*{0.0em}5: \quad \quad $\mathcal{PL}$ $ \leftarrow$ $\mathcal{PL}_{S} \leftarrow RPD (P_{S})$	\\
	\hspace*{0.0em}6: \quad \textbf{else}	\\
	\hspace*{0.0em}7: \quad\quad $\mathcal{PL}$ $ \leftarrow$ $\mathcal{PL}_{T} \leftarrow RPD(P_{T})$	\\
	\hspace*{0.0em}8: \quad \textbf{end if} 	\\
	\hspace*{0.0em}9: \quad\textbf{return} $\mathcal{PL}$  
	\vspace{0.3\baselineskip}
	\caption{Dynamic Competitive Selection Module}\label{alg1}
\end{algorithm}
\subsection{Dynamic Competitive Selection Module}
Algorithm \ref{alg1} details the proposed Dynamic Competitive Selection ($DCS$) module. Using predictions $\mathcal{P}_{S}$ and $\mathcal{P}_{T}$ from the student and teacher networks, respectively, we compute the partial cross-entropy loss $\mathcal{L}_{pCE}$ with respect to the scribble annotations $s$ using Equations \ref{equ1}, yielding $L_{pCE_{S}}$ and $L_{pCE_{T}}$. We assume that the network exhibiting superior performance within scribbles is likely to perform better across the full image. Based on this assumption, we select the better-performing prediction and convert it into pseudo labels using the $RPD$ module. Through this process, we obtain the final pseudo-label map $\mathcal{PL}$ for supervision. We assume that the network exhibiting superior performance within scribbles is likely to perform better across the full image. Based on this assumption, we select the better-performing prediction and convert it into pseudo labels using the $RPD$ module. Through this process, we obtain the final pseudo-label map $\mathcal{PL}$ for supervision. As shown in Fig.~\ref{acc_method_iters}, the results further confirm that integrating the $DCS$ mechanism effectively improves the accuracy of the generated pseudo labels, thereby validating our assumption.

\subsection{Total Loss Function}
The ScribbleVS framework is optimized by jointly learning from both the scribble annotations and the generated pseudo labels $\mathcal{PL}$. Supervision from scribble annotations is guided by the partial cross-entropy loss defined in Equation~\ref{equ1}, which is applied to the prediction of the student network. To incorporate supervision from the pseudo labels generated by the $RPD$ and $DCS$ modules, we introduce a regional pseudo-label loss $\mathcal{L}_{\mathcal{PL}}$, which measures the discrepancy between the student prediction $\mathcal{P}_S$ and the pseudo-label map $\mathcal{PL}$. This loss comprises two components: a partial cross-entropy loss and a partial Dice loss, both computed within the high-confidence region $\Omega$. The loss is defined as follows:

\begin{equation}
	\mathcal{L}_{_{\mathcal{PL}}CE}(p,\mathcal{PL})= - \sum_{i \in \mathrm{\Omega}} \sum_{k\in K}\mathcal{PL}[i,k] \cdot \log p[i,k]\,,
\end{equation}
\begin{equation}
	\mathcal{L}_{_{\mathcal{PL}}DC}(p,\mathcal{PL}) = 1-\frac{ 2 \cdot { \sum\limits_{i \in \mathrm{\Omega}}{ \sum\limits_{k \in K}}  p[i,k]\cdot \mathcal{PL}[i,k]} }{\sum\limits_{i \in \mathrm{\Omega}}{ \sum\limits_{k \in K}}p[i,k]^{2}+\sum\limits_{i \in \mathrm{\Omega}}{ \sum\limits_{k \in K}} \mathcal{PL}[i,k]^{2}}\,,
\end{equation}
\begin{equation}
	\mathcal{L}_{_{\mathcal{PL}}} =\frac{1}{2} (\mathcal{L}_{_{\mathcal{PL}}CE}(\mathcal{P}_{S},\mathcal{PL}) + \mathcal{L}_{_{\mathcal{PL}}DC}(\mathcal{P}_{S},\mathcal{PL}))\,,
	\label{equ8}
\end{equation}
where $K$ represents the set of class indices within the $\mathcal{PL}$, and $\mathrm{\Omega}$ denotes the regions within $\mathcal{PL}$ used for pseudo-label supervision. Both $\mathcal{PL}[i,k]$ and $p[i,k]$ represent the predicted probability that the $i$-th pixel belongs to class $k$, provided by the pseudo label and the prediction, respectively. 
The overall objective function is thus formulated as:

\begin{equation}
	\mathcal{L}_{total} = \mathcal{L}_{sup} + \lambda \cdot \mathcal{L}_{_{\mathcal{PL}}}\,,
\end{equation}
where $\mathcal{L}_{sup}$ and $\mathcal{L}_{_{\mathcal{PL}}}$ are defined in Equations \ref{equ1} and \ref{equ8}, respectively. To balance the contributions of the supervised loss $\mathcal{L}_{sup}$ and the pseudo-label loss $\mathcal{L}_{\mathcal{PL}}$ during training, we employ a time-dependent Gaussian warm-up function $\lambda(t)$, as proposed in \cite{laine2016temporal}. The warm-up schedule is given by:
\begin{align}
	\lambda (t) & = \left\{\begin{matrix}e^{(-5(1-\frac{t}{t_{warm}} )^2)} \quad t < t_{warm} \\
		1 \hspace*{6.0em} t \ge t_{warm} 
	\end{matrix}\right. \,,
\end{align}
where $t$ represents the current training step, and $t_{warm}$ denotes the maximum warming-up step.

\section{Experiments and Results}
\subsection{Dataset}
In this paper, we evaluated our method and compared it with state-of-the-art (SOTA) approaches using four publicly available datasets: the ACDC dataset and the MSCMRseg dataset for cardiac structure segmentation from 2D slices, WORD dataset for multi-organ segmentation from 3D volumes, and BraTS2020 dataset for more complex tumor segmentation task with irregular boundaries. 

\textbf{The ACDC dataset}\cite{8360453} contains cine-MRI scans from 100 patients acquired on two MRI scanners with varying magnetic strengths and resolutions. Manual annotations of the left ventricle (LV), myocardium (MYO) and right ventricle (RV), are provided. For weakly supervised learning, we adopt the scribble annotations from \cite{9389796}. 
Following the established protocol, the dataset was divided into 70 training, 15 validation, and 15 testing subjects. To maintain comparability with SOTA methods utilizing unpaired mask-based shape priors, the training data were further separated into 35 scribble-labeled and 35 mask-labeled subjects, noting that these masks were not used during training.

\textbf{The MSCMRseg dataset} \cite{10.1007/978-3-319-46723-8_67}\cite{8458220} comprises late gadolinium enhancement (LGE) MRI scans from 45 patients which diagnosed with cardiomyopathy. Compared to standard cardiac MRIs, these scans pose greater challenges for segmentation. Gold standard annotations for LV, MYO, and RV are provided. Scribble annotations for these structures were obtained from \cite{cyclemix}. Following them, we allocated 25 scans for training, 5 for validation, and reserved the remaining 15 scans for testing. 

\textbf{The WORD dataset} \cite{LUO2022102642} contains 150 abdominal CT scans  from distinct patients who underwent radiotherapy. Each 3D volume comprises 159-330 axial slices with a resolution of 512 $\times$ 512 pixels. The segmentation task involves seven abdominal organs: liver, spleen, left kidney, right kidney, stomach, gallbladder, and pancreas. Scribble-style annotations for both foreground and background were generated in the axial plane using an automated tool simulating manual input \cite{LUO2022102642}. The dataset was split into 100 scans for training, 20 for validation, and 30 for testing.

\textbf{The BraTS2020 dataset} \cite{menze2014multimodal}\cite{bakas2017advancing}\cite{bakas2018identifying} consists of 369 MRI volumes acquired in four modalities (T1, T1ce, T2, and FLAIR). They have a uniform resolution of 1.0 mm$^{3}$ with a matrix size of 240 $\times$ 240 $\times$ 155. The segmentation task involves two target structures: the peritumoral edema and the tumor core and. The dataset was randomly partitioned into 70\%, 15\%, and 15\% subsets for training, validation, and testing, respectively. Scribble annotations were synthetically produced following the procedure used in the WORD dataset \cite{han2024dmsps}.

\begin{table*}[t]
	\centering
	\caption{The performance (Dice Scores) on ACDC and MSCMRseg dataset of ScribbleVS compared with different SOTA method.}
	\label{table1}
	\begin{threeparttable}
			\begin{tabular}{l|llll|llll}\hline
				\multirow{2}{*}{Methods} & \multicolumn{4}{c|}{ACDC dataset} & \multicolumn{4}{c}{MSCMRseg dataset} \\ \cline{2-9} 
				& LV & MYO & RV & Mean & LV & MYO & RV & Mean \\ \hline
				Fullsup & 92.7$_{\pm7.7}$ & 89.7$_{\pm3.9}$ & 88.6$_{\pm8.6}$ & 90.3$_{\pm6.7}$ & 92.3$_{\pm4.2}$ & 83.3$_{\pm5.6}$ & 85.1$_{\pm6.5}$ & 86.9$_{\pm5.4}$ \\
				U-Net$_{pCE}$ \cite{Tang_2018_CVPR} & 59.1$_{\pm19.9}$ & 55.4$_{\pm10.3}$ & 65.0$_{\pm20.2}$ & 59.8$_{\pm16.8}$ & 70.9$_{\pm10.7}$ & 51.3$_{\pm12.9}$ & 45.4$_{\pm15.4}$ & 55.9$_{\pm13.0}$ \\
				MixUp \cite{mixup} & 80.3$_{\pm17.8}$ & 75.3$_{\pm11.6}$ & 76.7$_{\pm22.6}$ & 77.4$_{\pm17.3}$ & 61.0$_{\pm14.4}$ & 46.3$_{\pm14.7}$ & 37.8$_{\pm15.3}$ & 48.4$_{\pm14.8}$ \\
				Cutout \cite{cutout} & 83.2$_{\pm17.2}$ & 75.4$_{\pm13.8}$ & 81.2$_{\pm12.9}$ & 80.0$_{\pm14.6}$ & 45.9$_{\pm7.7}$ & 64.1$_{\pm13.6}$ & 69.7$_{\pm14.9}$ & 59.9$_{\pm12.1}$ \\
				CutMix \cite{cutmix} & 64.1$_{\pm35.9}$ & 73.4$_{\pm14.4}$ & 74.0$_{\pm21.6}$ & 70.5$_{\pm24.0}$ & 57.8$_{\pm6.3}$ & 62.2$_{\pm12.1}$ & 76.1$_{\pm10.5}$ & 65.4$_{\pm9.6}$ \\
				Puzzle Mix \cite{puzzlemix} & 66.3$_{\pm33.3}$ & 65.0$_{\pm23.1}$ & 55.9$_{\pm34.3}$ & 62.4$_{\pm30.2}$ & 6.1$_{\pm2.1}$ & 63.4$_{\pm8.4}$ & 2.8$_{\pm1.2}$ & 24.1$_{\pm3.9}$ \\
				Co-mixup \cite{comixup} & 62.2$_{\pm30.4}$ & 62.1$_{\pm21.4}$ & 70.2$_{\pm21.1}$ & 64.8$_{\pm24.3}$ & 35.6$_{\pm7.5}$ & 34.3$_{\pm6.7}$ & 5.3$_{\pm2.2}$ & 25.1$_{\pm5.5}$ \\
				CycleMix \cite{cyclemix} & 88.3$_{\pm9.5}$ & 79.8$_{\pm7.5}$ & 86.3$_{\pm7.3}$ & 84.8$_{\pm8.1}$ & 87.0$_{\pm6.1}$ & 73.9$_{\pm4.9}$ & 79.1$_{\pm7.2}$ & 80.0$_{\pm6.1}$ \\
				S2L \cite{Scribble2Label} & 87.2$_{\pm10.2}$ & 82.7$_{\pm4.6}$ & 77.4$_{\pm15.1}$ & 82.4$_{\pm10.0}$ & 80.3$_{\pm11.2}$ & 69.4$_{\pm8.9}$ & 63.9$_{\pm11.0}$ & 71.2$_{\pm10.3}$ \\
				GCL \cite{obukhov2019gated} & 91.8$_{\pm6.3}$ & 84.2$_{\pm4.3}$ & 87.0$_{\pm9.2}$ & 87.7$_{\pm6.6}$ & 92.1$_{\pm3.5}$ & 84.1$_{\pm3.5}$ & 86.8$_{\pm6.0}$ & 87.7$_{\pm4.3}$ \\
				ScribbleVC \cite{ScribbleVC} & 91.5$_{\pm5.5}$ & 86.7$_{\pm2.8}$ & 86.9$_{\pm7.8}$ & 88.4$_{\pm5.3}$ &91.6$_{\pm3.3}$ & 81.1$_{\pm3.3}$ & 86.1$_{\pm5.3}$ & 86.3$_{\pm4.0}$ \\
				DMSPS$_{stage1}$ \cite{han2024dmsps} & 92.7$_{\pm5.6}$ & 88.2$_{\pm3.2}$ & 87.6$_{\pm10.2}$ & 89.5$_{\pm6.3}$ & 91.8$_{\pm3.8}$ & 83.8$_{\pm3.2}$ & 87.5$_{\pm5.2}$ & 87.7$_{\pm4.0}$ \\
				DMSPS$_{stage2}$ \cite{han2024dmsps} & 92.7$_{\pm6.2}$ & 88.8$_{\pm2.8}$ & 88.0$_{\pm10.2}$ & 89.8$_{\pm6.2}$ & 90.0$_{\pm5.7}$ & 80.5$_{\pm3.6}$ & 83.1$_{\pm5.5}$ & 84.5$_{\pm5.0}$ \\
				ScribbleVS(Ours) & \textbf{92.9$_{\pm5.3}$} & $\textbf{89.4}^{\ast}_{\pm 2.7}$ & $\textbf{89.5}^{\ast}_{\pm 8.7}$& $\textbf{90.6}^{\ast}_{\pm 5.6}$ & $\textbf{93.6}^{\ast}_{\pm 2.5}$ & $\textbf{85.8}^{\ast}_{\pm 3.4}$ & $\textbf{86.6}^{\ast}_{\pm 7.7}$ & $\textbf{88.7}^{\ast}_{\pm 4.6}$ \\ \hline
			\end{tabular}
		\begin{tablenotes}
			\footnotesize
			\item Normality was assessed with the Shapiro-Wilk test. Paired $t$-tests were used for normally distributed data and Wilcoxon signed-rank tests otherwise. $^{\ast}$ denotes $p$-value$<$0.05 over DMSPS \cite{han2024dmsps}.
		\end{tablenotes}
	\end{threeparttable}
\end{table*}

\subsection{Implementation Details}
\textbf{2D segmentation:} For the ACDC and MSCMRseg datasets, we employed the U-Net \cite{unet} as the baseline model to ensure fair comparison. Slice intensities were normalized to [0,1], and data augmentation was performed with random rotation and flipping. Images were resized to 256$\times$256 for training, with batch size and $\tau$ set to 12 and 0.5, respectively. During testing, predictions were generated slice by slice and assembled into 3D volume. The Dice coefficient was used to assess performance. 

\textbf{3D segmentation:} For the WORD dataset, a fixed window/level of 400/50 was applied to emphasize abdominal organs, and Hounsfield Unit values were normalized to [0,1]. In BraTS2020, images were cropped to the non-zero bounding box, and all modalities were stacked to form a 4-channel input following~\cite{han2024dmsps}. For both datasets, we used 3D U-Net~\cite{cciccek20163d} with data augmentation by random cropping (80$\times$96$\times$96). The batch size was fixed at 8, and the value of $\tau$ was set to 0.65, respectively. Final segmentations were generated using a sliding-window approach. Evaluation was conducted with 3D Dice Similarity Coefficient and 95$\%$ Hausdorff Distance (HD95).

The proposed framework was implemented in PyTorch 1.12.0, using an Nvidia RTX 3090 GPU with 24GB of memory. All experiments were run for 60k iterations, with a warm-up period ($t_{\text{warm}}$) of 12k iterations. For model optimization, we employed SGD with initial learning rate of 0.01, with weight decay of $10^{-4}$ and momentum of 0.9 was used. The poly learning rate strategy was adopted to dynamically adjust the learning rate \cite{luo2021efficient}. 
\subsection{Performance Comparison with Other SOTA Methods}
To demonstrate the comprehensive segmentation performance of our method, we compared ScribbleVS with various SOTA methods.
\subsubsection{Baselines}
Across all tasks, we compared our method with existing approaches, including the partial cross-entropy loss (pCE), S2L \cite{Scribble2Label}, CycleMix \cite{cyclemix}, GCL \cite{obukhov2019gated}, and DMSPS \cite{han2024dmsps}. Additionally, for the ACDC and MSCMR datasets, we further evaluated our method against several data augmentation techniques, such as MixUp \cite{mixup}, Cutout \cite{cutout}, CutMix \cite{cutmix}, PuzzleMix \cite{puzzlemix}, and Co-mixup \cite{comixup}. In addition, we compare with ScribbleVC \cite{ScribbleVC} , which integrates scribble-based supervision with segmentation networks and class embedding modules. To ensure a comprehensive evaluation, we also included previously reported results \cite{9389796} on the ACDC dataset from U-Net with weighted partial cross-entropy (wpCE) \cite{9389796} and U-Net combined with conditional random fields (CRF) \cite{Zheng_2015_ICCV}.

\subsubsection{Challenging Benchmarks}
The methods mentioned above did not utilize additional unpaired segmentation masks during training. For a more rigorous evaluation, we additionally compared our method with four other approaches that incorporate shape priors from extra unpaired data: PostDAE \cite{postdae}, U-Net$_{D}$ \cite{9389796}, ACCL \cite{zhang2020accl}, and MAAG \cite{9389796}. Segmentation performances were compared on the ACDC dataset using the results reported in \cite{9389796}. Additionally, we conducted comparisons with established semi-supervised learning methods. To reflect annotation efforts similar to scribble-based methods, we used a training set of 7 labeled subjects and 28 unlabeled subjects. We evaluated our approach against four widely used semi-supervised segmentation methods: MT \cite{tarvainen2017mean}, UAMT \cite{yu2019uncertainty}, SLC-Net \cite{slcnet}, and URPC \cite{luo2022semi}.

\begin{figure*}[t]
	\centering
	\includegraphics[width=1\textwidth]{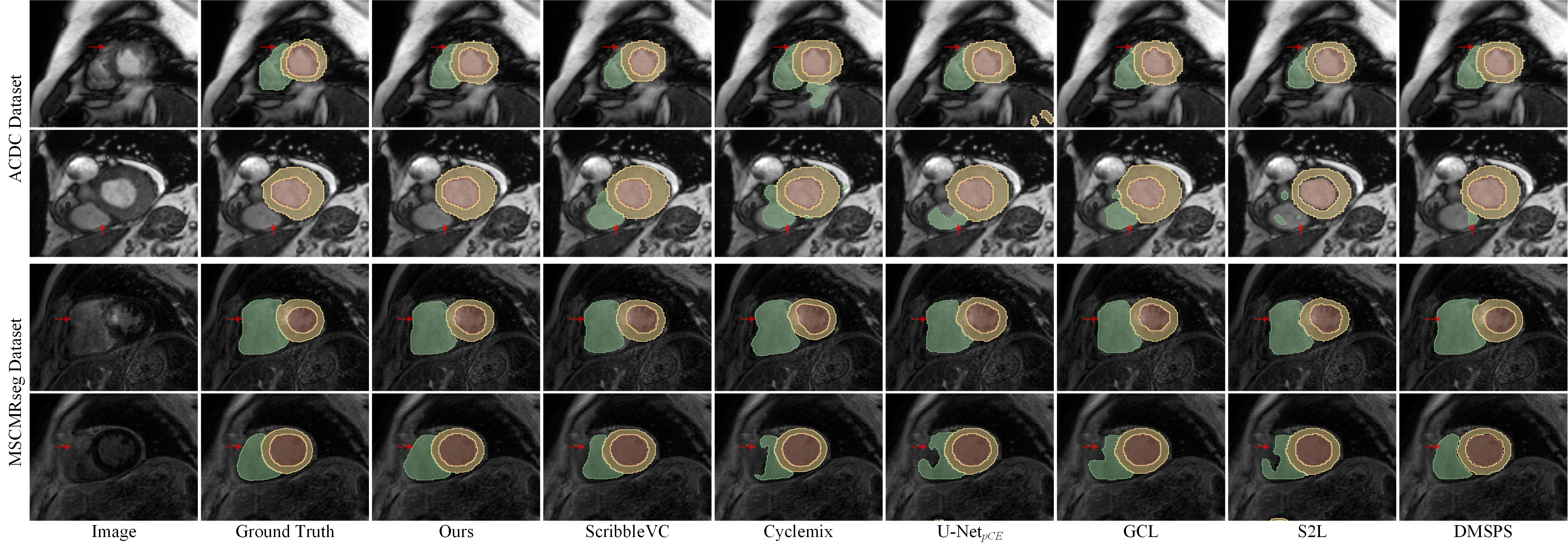}		
	\caption{Visualization with other methods on the ACDC and MSCMRseg datasets. The models are trained with scribble annotations.}
	\label{figure20}
\end{figure*}

\subsection{Result on ACDC and MSCMR dataset}

Table \ref{table1} presents the segmentation performance on the ACDC and MSCMRseg datasets. Our proposed ScribbleVS model, which utilizes scribble annotations, consistently outperforms various training strategies, model architectures, and data augmentation techniques. 
Specifically, our method outperforms DMSPS$_{stage2}$. On the ACDC dataset, it achieves a 0.8$\%$ improvement over DMSPS$_{stage2}$ (90.6$\%$ vs 89.8$\%$), while on the MSCMRseg dataset, it achieves a 1.0$\%$ improvement compared to DMSPS$_{stage1}$ (88.7$\%$ vs 87.7$\%$). 
Notably, the decline of DMSPS$_{stage2}$ on the MSCMRseg dataset indicates that erroneous expansions in the annotation refinement process likely introduced additional noise, which undermined the accuracy achieved at $stage1$.
These results underscore the effectiveness of our method. In contrast, U-Net$_{pCE}$ exhibited weaker performance, achieving Dice scores of 76.6$\%$ and 68.9$\%$ on ACDC and MSCMRseg, respectively. Particularly, significant performance degradation was noted on the MSCMRseg dataset, underscores the adaptability of our framework in handling datasets with more segmentation complexities.
Fig. \ref{figure20} provides qualitative comparisons between ScribbleVS and other SOTA methods on the ACDC and MSCMRseg datasets. As shown, our approach produces segmentation results that are visually more consistent with the ground truth, offering a clear and intuitive demonstration of its superiority.
\begin{table}[t]
	\centering
	\caption{Comparison of ScribbleVS and state-of-the-art weakly supervised methods on the ACDC dataset (Dice scores), using results from \cite{9389796}. }
	\label{table2}
	\resizebox{0.48\textwidth}{!} {
		\begin{tabular}{lcccc}
			\hline
			\multicolumn{1}{l|}{Methods}                                                 & LV            & MYO           & RV            & Avg           \\ \hline
			\multicolumn{5}{l}{35 scribbles}                                                                                                             \\ \hline
			\multicolumn{1}{l|}{U-Net$_{pCE}$ \cite{Tang_2018_CVPR}}  & 84.2$_{\pm7}$          & 76.4$_{\pm6}$          & 69.3$_{\pm11}$          & 76.6$_{\pm8}$          \\
			\multicolumn{1}{l|}{U-Net$_{wpCE}$ \cite{9389796}}          & 78.4$_{\pm9}$          & 67.5$_{\pm6}$          & 56.3$_{\pm13}$          & 67.4$_{\pm9}$          \\
			\multicolumn{1}{l|}{U-Net$_{CRF}$ \cite{Zheng_2015_ICCV}} & 76.6$_{\pm9}$          & 66.1$_{\pm6}$          & 59.0$_{\pm14}$          & 67.2$_{\pm10}$          \\
			\multicolumn{1}{l|}{ScribbleVS(Ours)}                                        & \textbf{92.9$_{\pm5.3}$} & $\textbf{89.4}^{\ast}_{\pm 2.7}$ & $\textbf{89.5}^{\ast}_{\pm 8.7}$& $\textbf{90.6}_{\pm 5.6}$ \\ \hline
			\multicolumn{5}{l}{35 scribbles + 35 unpaired mask}                                                                                          \\ \hline
			\multicolumn{1}{l|}{U-Net$_{D}$ \cite{9389796}}             & 75.3$_{\pm9}$          & 59.7$_{\pm8}$          & 40.4$_{\pm15}$          & 58.5$_{\pm11}$          \\
			\multicolumn{1}{l|}{PostDAE \cite{postdae}}                 & 80.6$_{\pm7}$          & 66.7$_{\pm7}$          & 55.6$_{\pm12}$          & 67.5$_{\pm9}$          \\
			\multicolumn{1}{l|}{ACCL \cite{zhang2020accl}}              & 87.8$_{\pm6}$          & 79.7$_{\pm5}$          & 73.5$_{\pm10}$          & 80.3$_{\pm7}$          \\
			\multicolumn{1}{l|}{MAAG \cite{9389796}}                    & 87.9$_{\pm5}$          & 81.7$_{\pm5}$          & 75.2$_{\pm12}$          & 81.6$_{\pm7}$          \\ \hline
			\multicolumn{5}{l}{7 paired mask and 28 unlabeled}                                                                                           \\ \hline
			\multicolumn{1}{l|}{MT \cite{tarvainen2017mean}}            & 85.8$_{\pm15.7}$          & 78.6$_{\pm14.5}$          & 67.8$_{\pm26.4}$          & 77.4$_{\pm18.9}$          \\
			\multicolumn{1}{l|}{UAMT \cite{yu2019uncertainty}}         & 87.2$_{\pm12.2}$          & 80.4$_{\pm11.2}$          & 71.7$_{\pm25.9}$          & 79.8$_{\pm16.4}$          \\
			\multicolumn{1}{l|}{SLC-Net \cite{slcnet}}                  & 87.3$_{\pm9.7}$          & 80.7$_{\pm9.8}$          & 74.5$_{\pm20.6}$          &80.8$_{\pm13.4}$          \\
			\multicolumn{1}{l|}{URPC \cite{luo2022semi}}                & 84.5$_{\pm14.2}$          & 76.7$_{\pm14.9}$          & 71.4$_{\pm25.5}$          & 77.5$_{\pm18.1}$          \\ \hline
			\end{tabular} 
		}
\end{table}

\textbf{Comparison with other weakly supervised methods:} Table~\ref{table2} summarizes the segmentation performance on the ACDC dataset. The previous best-performing method, MAAG~\cite{9389796}, incorporated multi-scale GANs and unpaired masks from 35 additional subjects, achieving a Dice score of 81.6$\%$. In contrast, ScribbleVS attains an average Dice score of 90.6$\%$ without relying on such extra masks. Notably, for the RV, a structure with high anatomical variability, ScribbleVS improves performance by 14.3$\%$ over MAAG (89.5$\%$ vs. 75.2$\%$). These results indicate that GAN-based approaches trained on limited data, even with auxiliary masks, struggle to capture sufficient shape priors. In a semi-supervised setting with 7 labeled and 28 unlabeled subjects, SLC-Net achieves a Dice score of 80.8$\%$, still far below ScribbleVS. This demonstrates that, under comparable annotation costs, our method delivers substantially superior segmentation performance. Moreover, ScribbleVS consistently outperforms other scribble-supervised methods, achieving up to 14.0$\%$ average improvement.

\begin{table*}[t]
	\centering
	\caption{Quantitative comparison between our method ScribbleVS and existing weakly supervised methods on WORD dataset. }
	\label{table_WORD}
	\begin{threeparttable}
	\resizebox{0.85\textwidth}{!} {
		\begin{tabular}{l@{\hskip 2.8pt}ll@{\hskip 2.8pt}l@{\hskip 2.8pt}l@{\hskip 2.8pt}l@{\hskip 2.8pt}l@{\hskip 2.8pt}l@{\hskip 2.8pt}l@{\hskip 2.8pt}l@{\hskip 2.8pt}}
			\hline
			Metric &Method & Liver & Spleen & Kidney(L) & Kidney(R) & Stomach & Gallbladder & Pancreas & Mean \\
			\hline
			\multirow{8}{*}{\makecell{Dice\\(\%)}}       
			& FullySup & 96.4$_{\pm 0.7}$ & 95.4$_{\pm 1.6}$ & 95.0$_{\pm 1.6}$ & 95.3$_{\pm 1.3}$ & 90.1$_{\pm 4.4}$ & 75.3$_{\pm 13.2}$ & 80.9$_{\pm 7.7}$ & 89.8$_{\pm 4.4}$ \\
			& 3D UNet$_{pCE}$ \cite{Tang_2018_CVPR} & 86.5$_{\pm 3.6}$ & 86.8$_{\pm 5.8}$ & 86.2$_{\pm 4.5}$ & 89.4$_{\pm 3.0}$ & 61.1$_{\pm 12.2}$ & 56.6$_{\pm 20.1}$ & 61.5$_{\pm 10.4}$ & 75.5$_{\pm 8.5}$ \\
			& S2L \cite{Scribble2Label} & 68.4$_{\pm 3.5}$ & 90.0$_{\pm 5.5}$ & 88.1$_{\pm 4.0}$ & 89.9$_{\pm 4.0}$ & 81.4$_{\pm 6.1}$ & 61.0$_{\pm 19.6}$ & 70.0$_{\pm 8.5}$ & 78.4$_{\pm 7.3}$ \\
			& CycleMix\cite{cyclemix} & 94.2$_{\pm 0.9}$ & 87.3$_{\pm 5.8}$ & 90.7$_{\pm 1.8}$ & 90.9$_{\pm 3.8}$ & 82.3$_{\pm 5.8}$ & 66.0$_{\pm 18.2}$ & 73.8$_{\pm 7.4}$ & 83.6$_{\pm 6.2}$ \\
			& GCL \cite{obukhov2019gated} & 93.5$_{\pm 1.1}$ & 93.3$_{\pm 2.5}$ & 90.7$_{\pm 3.2}$ & 92.5$_{\pm 3.2}$ & 83.5$_{\pm 4.6}$ & 69.6$_{\pm 15.0}$ & 72.7$_{\pm 8.6}$ & 85.1$_{\pm 5.5}$ \\
			& DMSPS$_{stage1}$ \cite{han2024dmsps} & {94.4$_{\pm 0.9}$} & {92.2$_{\pm 2.7}$} & {92.4$_{\pm 2.4}$} & {93.0$_{\pm 1.7}$} & {87.9$_{\pm 3.8}$} & {71.1$_{\pm 15.4}$} & {76.5$_{\pm 7.8}$} & {86.8$_{\pm 4.8}$} \\
			& DMSPS$_{stage2}$ \cite{han2024dmsps} & \textbf{94.8$_{\pm 0.7}$} & {93.4$_{\pm 2.1}$} & \textbf{93.3$_{\pm 2.0}$} & {93.6$_{\pm 1.5}$} & {88.4$_{\pm 3.7}$} & {72.3$_{\pm 14.8}$} & {77.1$_{\pm 7.3}$} & {87.6$_{\pm 4.6}$} \\
			& Ours & {94.4$_{\pm 1.3}$} & \textbf{93.5$_{\pm 2.1}$} & {93.2$_{\pm 1.5}$} & \textbf{93.8$_{\pm 1.2}$}  & \textbf{88.7$_{\pm 3.5}$} &  $\textbf{74.5}^{\ast}_{\pm 10.8}$&  $\textbf{78.3}^{\ast}_{\pm 6.7}$& $\textbf{88.1}^{\ast}_{\pm 2.0}$ \\
			\hline
			\multirow{8}{*}{\makecell{HD95\\(mm)}}       
			& FullySup & 3.5$_{\pm 1.4}$ & 3.0$_{\pm 3.3}$ & 2.8$_{\pm 1.1}$ & 2.5$_{\pm 0.7}$ & 10.0$_{\pm 8.4}$ & 10.5$_{\pm 15.8}$ & 7.7$_{\pm 6.9}$ & 5.7$_{\pm 5.4}$ \\
			& 3D UNet$_{pCE}$ \cite{Tang_2018_CVPR} & 65.0$_{\pm 33.4}$ & 59.6$_{\pm 46.2}$ & 63.5$_{\pm 46.8}$ & 56.1$_{\pm 50.3}$ & 130.1$_{\pm 30.5}$ & 69.2$_{\pm 37.0}$ & 99.4$_{\pm 13.3}$ & 77.6$_{\pm 36.8}$ \\
			& S2L \cite{Scribble2Label} & 48.4$_{\pm 14.4}$ & 6.7$_{\pm 6.0}$ & 6.5$_{\pm 1.9}$ & 6.0$_{\pm 2.4}$ & 38.8$_{\pm 41.9}$ & 26.5$_{\pm 29.3}$ & 34.6$_{\pm 29.0}$ & 23.9$_{\pm 17.9}$ \\
			& CycleMix\cite{cyclemix} & 6.3$_{\pm 5.2}$ & 12.6$_{\pm 13.0}$ & 4.9$_{\pm 3.1}$ & 4.2$_{\pm 3.1}$ & 16.3$_{\pm 13.9}$ & 17.9$_{\pm 14.4}$ & 10.3$_{\pm 7.2}$ & 10.4$_{\pm 8.6}$ \\
			& GCL \cite{obukhov2019gated} & 7.3$_{\pm 3.1}$ & 7.2$_{\pm 18.1}$ & 7.3$_{\pm 2.2}$ & 4.4$_{\pm 2.0}$ & 14.6$_{\pm 8.6}$ & 11.8$_{\pm 16.4}$ & 16.7$_{\pm 23.7}$ & 9.9$_{\pm 10.6}$ \\
			& DMSPS$_{stage1}$ \cite{han2024dmsps} & {5.6$_{\pm 2.6}$} & {9.4$_{\pm 2.8}$} & {4.4$_{\pm 2.0}$} & {3.6$_{\pm 1.1}$} & {10.0$_{\pm 6.3}$} & {17.1$_{\pm 22.9}$} & {12.0$_{\pm 16.3}$} & {8.8$_{\pm 11.3}$} \\
			& DMSPS$_{stage2}$ \cite{han2024dmsps} & \textbf{5.0$_{\pm 1.8}$} & \textbf{3.5$_{\pm 1.6}$} & \textbf{3.7$_{\pm 1.6}$} & {3.3$_{\pm 0.8}$} & {9.9$_{\pm 7.1}$} & {9.0$_{\pm 13.8}$} & {9.5$_{\pm 7.7}$} & {6.3$_{\pm 4.9}$} \\
			& Ours & \textbf{5.0$_{\pm 1.9}$} & {3.7$_{\pm 2.2}$} & {3.8$_{\pm 1.3}$} & \textbf{3.2$_{\pm 0.7}$}  & \textbf{8.8$_{\pm 4.8}$} & $\textbf{6.3}^{\ast}_{\pm 6.1}$ & $\textbf{8.6}^{\ast}_{\pm 8.0}$ & $\textbf{5.6}^{\ast}_{\pm 1.8}$\\
			\hline 
		\end{tabular}
		}
	\end{threeparttable}
\end{table*}

\subsection{Results on the WORD dataset}

\begin{figure}[t]
	\centering
	\includegraphics[width=0.49\textwidth]{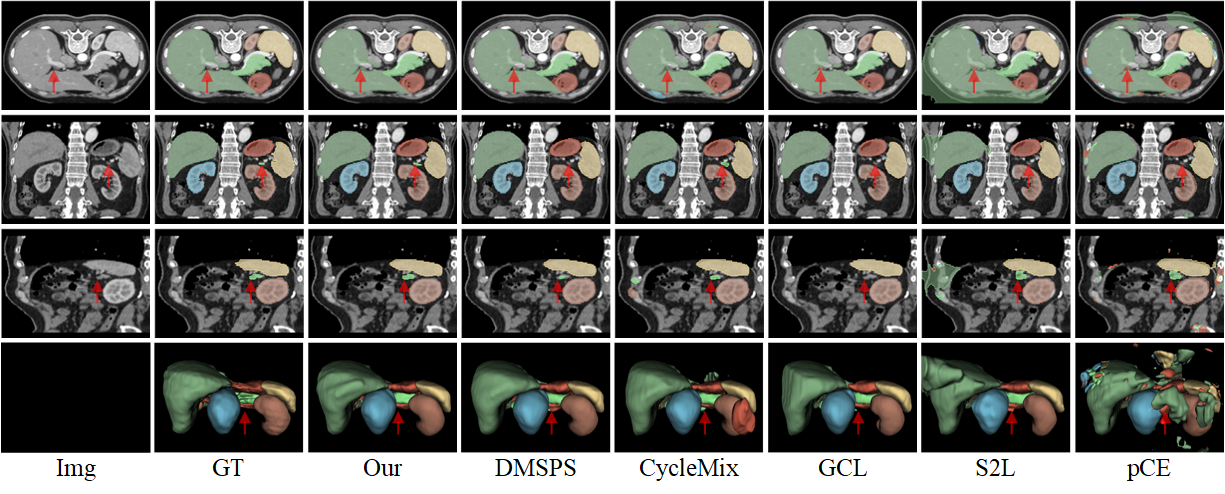}		
	\caption{Visual comparison between our method and the other weakly supervised methods on the WORD dataset. From top to botton, the visualizations correspond to axial, coronal, sagittal, and 3D views.}
	\label{figure_word_seg}
\end{figure}

Our method was further examined using the WORD dataset, representing a more demanding 3D multi-organ segmentation task. Quantitative comparisons in terms of Dice and HD95 metrics are summarized in Table \ref{table_WORD}. The baseline pCE method achieved an average Dice of 75.5$\%$, while the two-stage framework DMSPS attained the highest Dice (87.6$\%$) among existing approaches. Our proposed ScribbleVS surpassed the DMSPS, achieving an average Dice of 88.1$\%$ and an HD95 of 5.6 mm. ScribbleVS demonstrated consistent improvements across most organs, particularly excelling in anatomically complex or smaller structures such as the gallbladder and pancreas.
Fig. \ref{figure_word_seg} visually compares segmentation results from different methods. While the $pCE$ method produced numerous false positives and other methods yielded noisy segmentation outputs, our ScribbleVS delivered significantly cleaner and more accurate segmentations, effectively mitigating both over- and under-segmentation errors.

\begin{table}[t]
	\centering
	\caption{Quantitative comparison between our ScribbleVS and existing weakly supervised methods on BraTS2020 dataset.}
	\label{table_bra2020}
	\begin{threeparttable}
    \resizebox{0.5\textwidth}{!} {
	\begin{tabular}{l|ll|ll|ll}
        \hline
        \multirow{2}{*}{Method}                                  & \multicolumn{2}{l|}{Tumor}                                           & \multicolumn{2}{l|}{Edema}                                           & \multicolumn{2}{l}{Mean}                                                                                          \\ \cline{2-7} 
         & Dice(\%)  & HD95 & Dice(\%) & HD95 & Dice(\%) & HD95                             \\ \hline
        FS (upper bound)                                         & 82.0$_{\pm21.5}$                  & 9.1$_{\pm20.2}$                  & 78.6$_{\pm13.5}$                  & 6.5$_{\pm13.4}$                  & 80.3$_{\pm17.5}$                                                               & 7.8$_{\pm16.8}$                  \\
        3D UNet$_{pCE}$ \cite{Tang_2018_CVPR} & 56.8$_{\pm24.0}$                  & 74.1$_{\pm36.8}$                 & 50.5$_{\pm18.6}$                  & 67.9$_{\pm20.6}$                 & 53.6$_{\pm21.3}$                                                               & 71.0$_{\pm28.7}$                 \\
        USTM \cite{liu2022weakly}               & 63.9$_{\pm24.2}$                  & 69.2$_{\pm44.7}$                 & 46.9$_{\pm18.1}$                  & 84.1$_{\pm17.7}$                 & 55.4$_{\pm21.2}$                                                               & 76.6$_{\pm31.2}$                 \\
        S2L \cite{Scribble2Label}               & 75.8$_{\pm22.5}$                  & 17.8$_{\pm28.8}$                 & 61.5$_{\pm12.2}$                  & 25.3$_{\pm27.1}$                 & 68.6$_{\pm17.3}$                                                               & 21.6$_{\pm27.9}$                 \\
        CycleMix \cite{cyclemix}                & 77.7$_{\pm17.7}$                  & 28.8$_{\pm34.2}$                 & 64.2$_{\pm14.4}$                  & 40.5$_{\pm28.7}$                 & 71.0$_{\pm16.0}$                                                               & 34.7$_{\pm31.5}$                 \\
        GCL \cite{obukhov2019gated}             & 77.2$_{\pm21.9}$                  & 18.3$_{\pm29.7}$                 & 67.9$_{\pm12.7}$                  & 29.8$_{\pm33.7}$                 & 72.6$_{\pm17.3}$                                                               & 24.0$_{\pm31.7}$                 \\
        DMSPS$_{stage1}$ \cite{han2024dmsps}    & 78.9$_{\pm21.8}$                    & 10.8$_{\pm20.6}$                   & 69.7$_{\pm12.4}$                    & 10.7$_{\pm17.7}$                   & 74.3$_{\pm17.1}$                                                                 & 10.8$_{\pm19.2}$                   \\
        DMSPS$_{stage2}$ \cite{han2024dmsps}    & 79.4$_{\pm21.2}$                    & 8.1$_{\pm16.1}$                    & 73.7$_{\pm12.0}$                    & 7.6$_{\pm13.1}$                    & 76.5$_{\pm16.6}$                                                                 & 7.9$_{\pm14.6}$                    \\
        Ours                                                     & $\textbf{82.1}^{\ast}_{\pm 18.6}$ & $\textbf{7.1}^{\ast}_{\pm 14.5}$ & $\textbf{75.2}^{\ast}_{\pm 11.1}$ & $\textbf{6.9}^{\ast}_{\pm 13.2}$ & $\textbf{78.6}^{\ast}_{\pm 14.8}$ & $\textbf{7.0}^{\ast}_{\pm 13.8}$ \\ \hline
        \end{tabular}  }
	\end{threeparttable}
\end{table}

\subsection{Results on the BraTS2020 dataset}
\begin{figure}[t]
	\centering
	\includegraphics[width=0.5\textwidth]{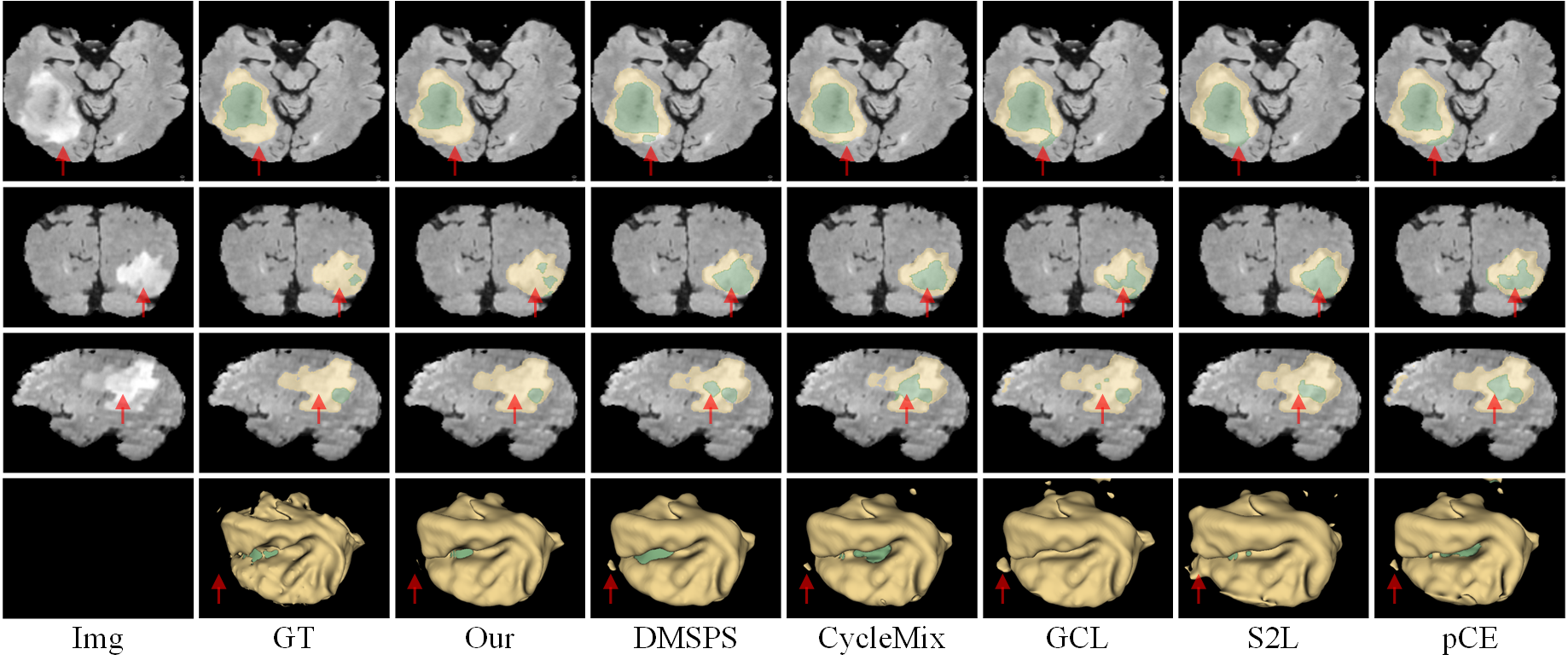}		
	\caption{Visual comparison between our method and the other weakly supervised methods on the BraTS2020 dataset. From top to botton, the visualizations correspond to axial, coronal, sagittal, and 3D views.}
	\label{figure_bra_seg}
\end{figure}

\color{black}
ScribbleVS was further evaluated on the BraTS2020 dataset, which involves the challenging task of brain tumor segmentation. Fig. \ref{figure_bra_seg} provides visual comparisons, showing that our method produced segmentation results closer to ground truth. In contrast, other methods exhibited considerable noise in their predictions, particularly in regions affected by edema. As reported in Table \ref{table_bra2020}, the baseline pCE method achieved a Dice score of 53.6$\%$ and an average HD95 of 71.0 mm. Among existing methods, the DMSPS$_{stage2}$ attained the best Dice of 76.5$\%$. Our method outperformed DMSPS, reaching an average Dice of 78.6$\%$ and an HD95 of 7.0 mm. These results demonstrate the generalization capability of our proposed framework.
\section{Ablation Study}
\color{black}
\label{section_abu}
\subsection{Evolution of Pseudo Labels via {$RPD$}} 
\begin{figure}[t]
	\centering
	\includegraphics[width=0.48\textwidth]{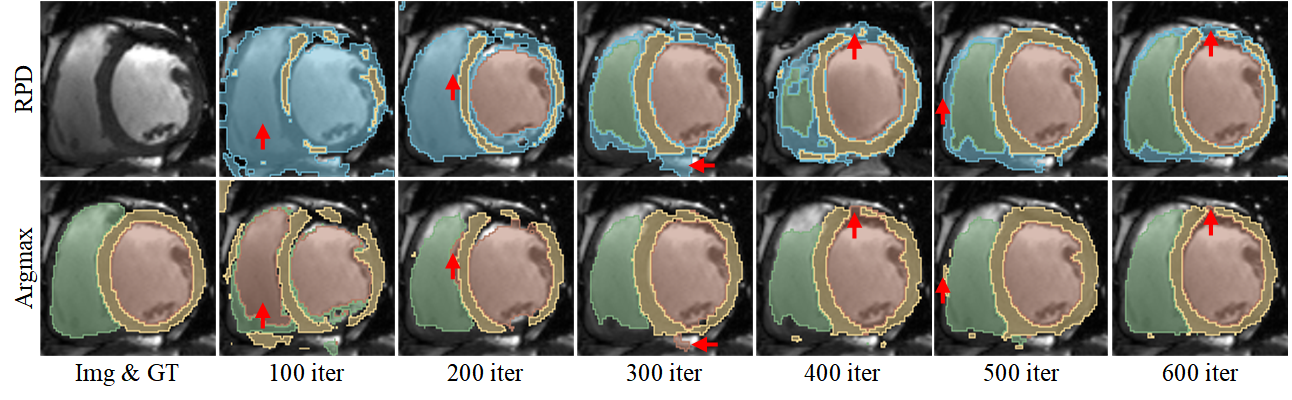}		
	\caption{Visualization of the evolution of pseudo labels generated by both the $RPD$ module and the argmax processing. Regions marked in \textcolor{mycolor_light_blue}{\rule{0.2cm}{0.2cm}} are excluded from loss calculation through Equ. \eqref{equ8}. Regions with clear differences are highlighted with red arrows.}
	\label{figure4}
\end{figure}

Fig. \ref{figure4} illustrates the evolution of pseudo-labels generated by the proposed $RPD$ module and argmax operation during training. During the early training stage (e.g., after 100 iterations), most regions selected by $RPD$ are marked as uncertain and thus excluded from loss computation. In contrast, argmax-based labels introduce substantial noise. As training proceeds, $RPD$ progressively reduces the number of uncertain regions, and the model's ability to distinguish boundaries between segmentation categories improves. This progression reflects enhanced confidence and accuracy in target delineation. Notably, $RPD$ avoids using low-confidence predictions, particularly in ambiguous boundary areas (highlighted by red arrows), thus reducing noise and promoting model robustness.
\subsection{Impact of Module Integration}

\begin{figure}[t]
	\centering
	\includegraphics[width=0.5\textwidth]{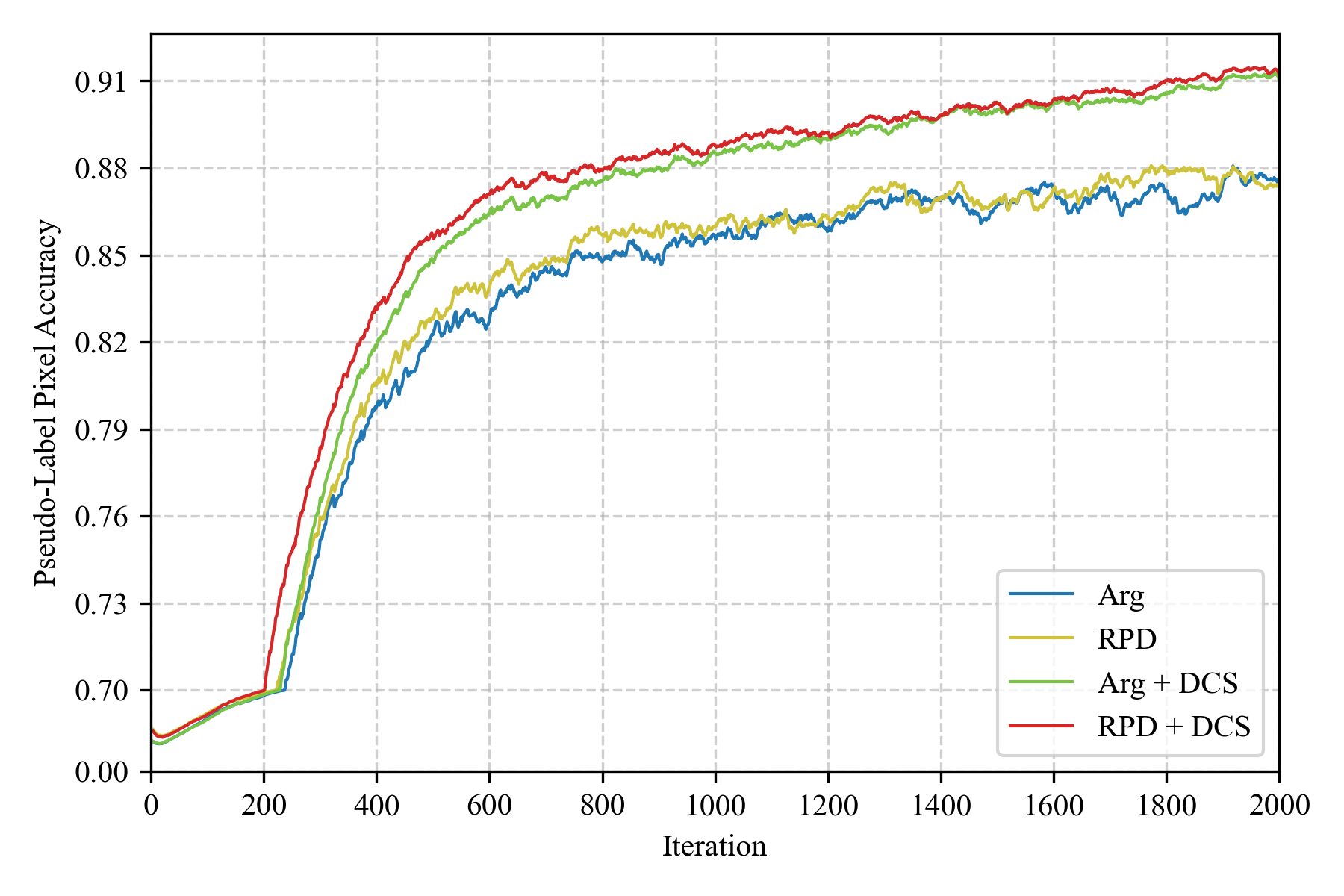}		
	\caption{Effect of $RPD$ and $DCS$ on pseudo-label accuracy.}
	\label{acc_method_iters}
\end{figure}

We conducted an ablation study to evaluate the contributions of individual components within ScribbleVS, including the $RPD$ module and the $DCS$ mechanism. Results in Table \ref{table32} show that using only $RPD$ module achieved a Dice score of 89.5$\%$, underscoring its effectiveness. When integrated into the mean teacher framework alongside $DCS$, the model exhibited further improvement, achieving a Dice score of 90.6$\%$. This highlights the effectiveness of employing a dynamic competitive selection strategy, which enhances network collaboration and contributes to greater model stability and performance. 
In addition, we further examined the accuracy of the pseudo-labels generated under different module combinations, as illustrated in Fig.~\ref{acc_method_iters}. The results reveal that $RPD$ substantially improves pseudo-label accuracy compared with the baseline, while its integration with $DCS$ leads to further gains, thereby providing more reliable supervision and explaining the observed performance improvements.

\begin{table}[t]
	\centering
	\caption{Ablation study about the combination of modules on the ACDC dataset. The Arg represents the pseudo labels generated by the Argmax process directly. }
	\label{table32}
	\resizebox{0.45\textwidth}{!} {
		\begin{tabular}{ccc|llll}
			\hline
			Arg & $RPD$ & $DCS$ & LV    & MYO   & RV    & Avg   \\ \hline
			\checkmark   &     &     & 81.5$_{\pm14.3}$& 80.4$_{\pm5.0}$ & 79.7$_{\pm17.6}$ & 80.5$_{\pm12.3}$ \\
			& \checkmark   &     & \textbf{93.0}$_{\pm4.5}$  & 87.7$_{\pm2.9}$ & 87.7$_{\pm8.9}$ & 89.5$_{\pm5.4}$ \\
			\checkmark   &     & \checkmark   & 92.9$_{\pm6.0}$ & 88.0$_{\pm2.8}$ & 86.6$_{\pm10.1}$ & 89.2$_{\pm6.3}$ \\
			& \checkmark   & \checkmark   & {92.9$_{\pm5.3}$} & $\textbf{89.4}^{\ast}_{\pm 2.7}$ & $\textbf{89.5}^{\ast}_{\pm 8.7}$& $\textbf{90.6}_{\pm 5.6}$ \\ \hline
	\end{tabular} 
}
\end{table}

\subsection{Effect of Confidence Threshold $\tau$}

\begin{figure}[!t]
	\centering
	\includegraphics[width=0.45\textwidth]{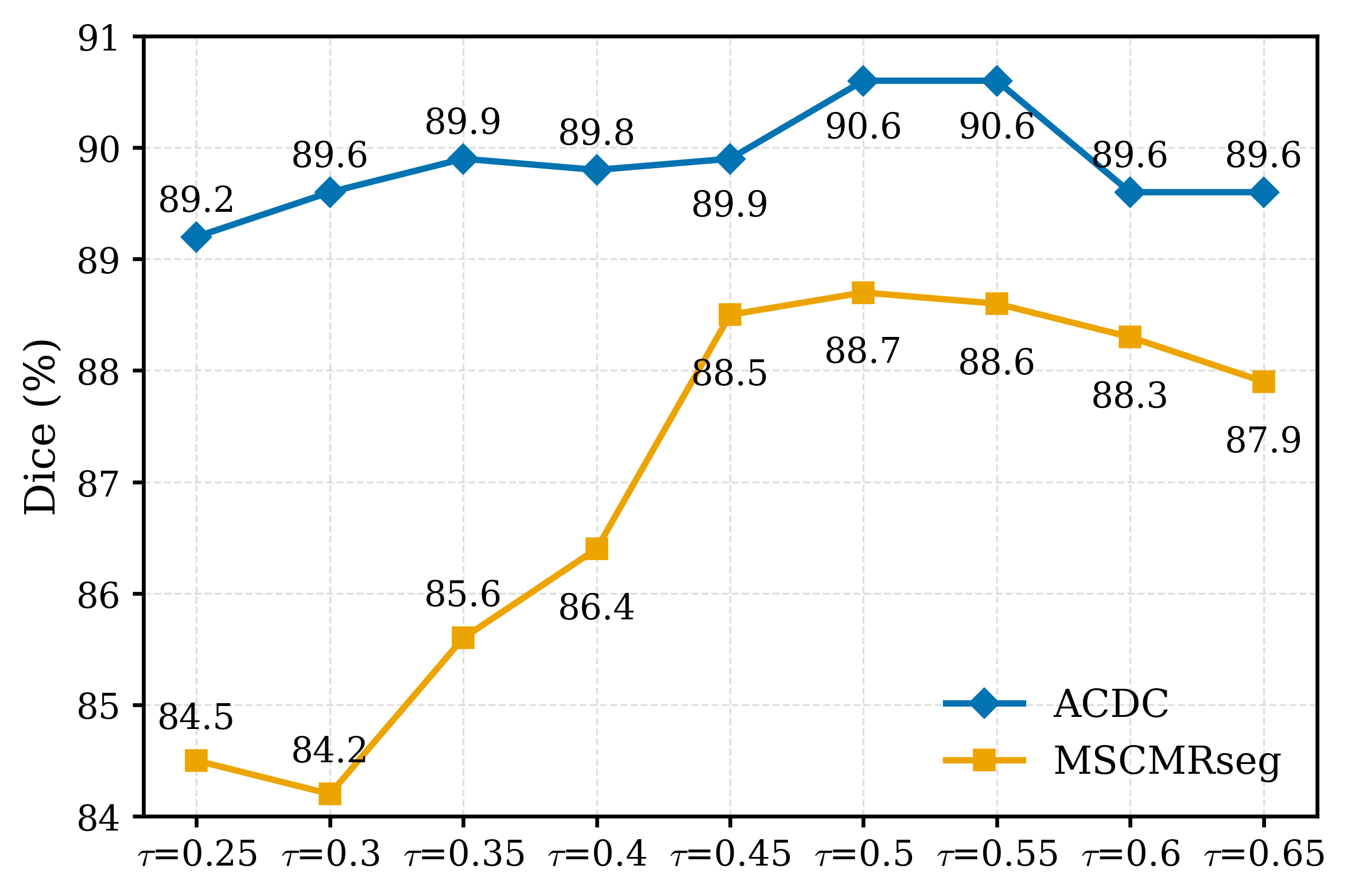}		
	\caption{Ablation study: the performance of ScribbleVS with different $\tau$ on the ACDC dataset and MSCMRseg dataset.}
	\label{figure5}
\end{figure}

The $RPD$ module employs a confidence threshold $\tau$ to filter low-confidence predictions during pseudo-label generation. Fig. \ref{figure5} presents the performance across varying $\tau$ values on the ACDC and MSCMRseg datasets. A lower $\tau$ includes more regions to be considered but more likely to introduce noise, while a higher $\tau$ improves label quality at the cost of reduced coverage.
On the ACDC dataset, optimal Dice scores of 90.6$\%$ were achieved at $\tau=0.5$ and $\tau=0.55$. For MSCMRseg dataset, the model attained the best performance with a Dice score of 88.7$\%$ at $\tau=0.5$, slightly outperforming the results at $\tau=0.55$. Consequently, $\tau=0.5$ was chosen for experiments on both datasets. 
Importantly, both datasets involve four-class segmentation. When $\tau$ is set to 0.25, the behavior of the $RPD$ module becomes equivalent to argmax operation. Moreover, the experimental results clearly demonstrate the superiority of $RPD$ over the argmax operation.  

\subsection{Data Sensitivity Experiments}
\begin{figure}[t]
	\centering
	\includegraphics[width=0.48\textwidth]{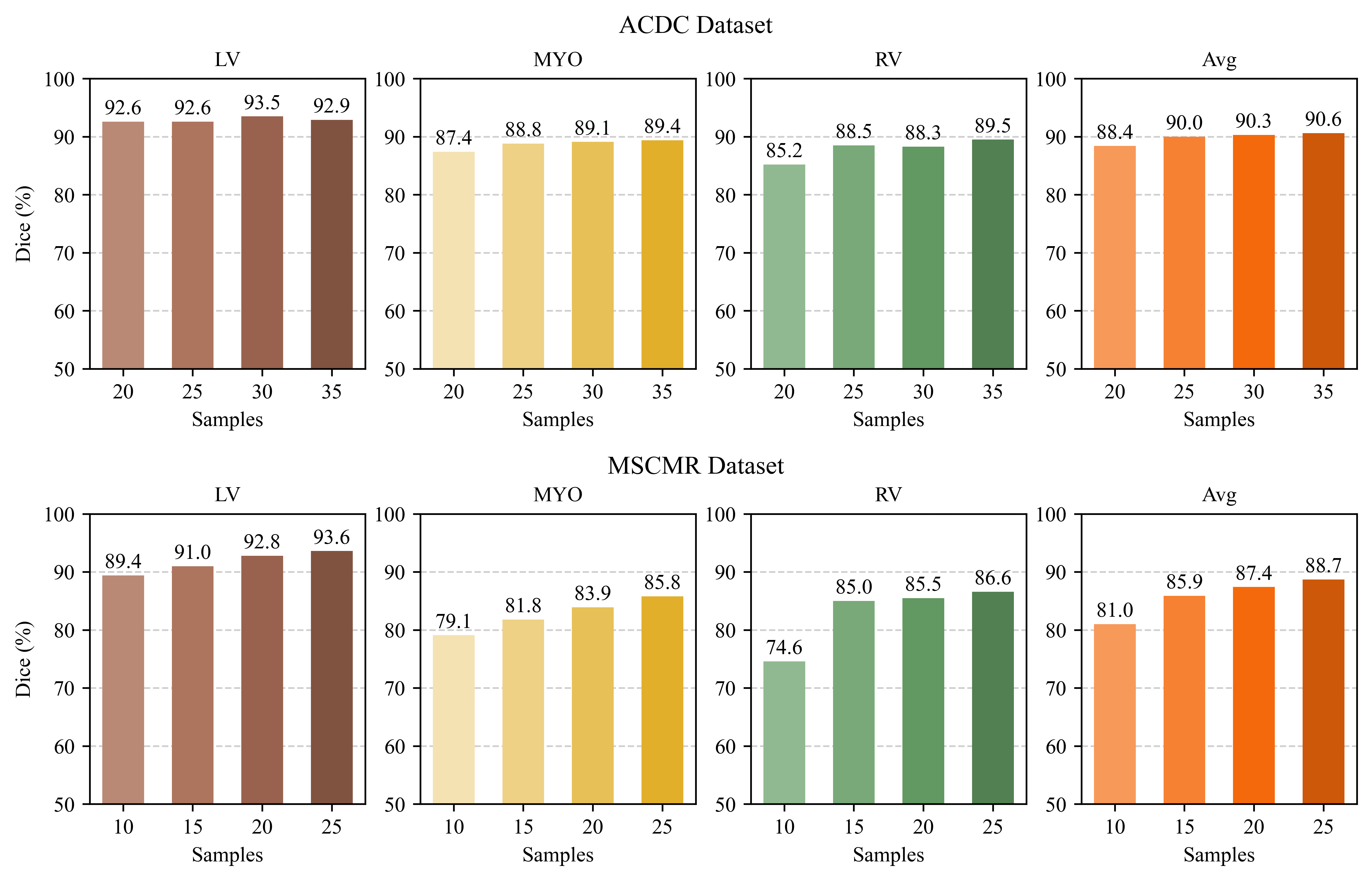}		
	\caption{Data sensitivity study: visualization of performance of ScribbleVS with the different numbers of scribbles for training.}
	\label{figure_result_num}
\end{figure}

\color{black}
We conducted a data sensitivity analysis to evaluate the performance of ScribbleVS with different amounts of scribble-annotated data. As shown in Fig. \ref{figure_result_num}, the model's performance improves steadily as the number of annotated samples increases. Remarkably, ScribbleVS reaches an average Dice score of 88.4$\%$ with just 20 annotated cases in the ACDC dataset and 85.9$\%$ with 15 cases in the MSCMRseg dataset. Furthermore, when trained with 25 annotated samples, the model maintains its strong performance with a Dice score of 90.0$\%$ in the ACDC dataset.These results highlight ScribbleVS’s robustness when trained with limited supervision. This resilience makes ScribbleVS well-suited for use in scenarios with limited annotated data.

\section{Discussion}
\begin{figure}[t]
	\centering
	\includegraphics[width=0.48\textwidth]{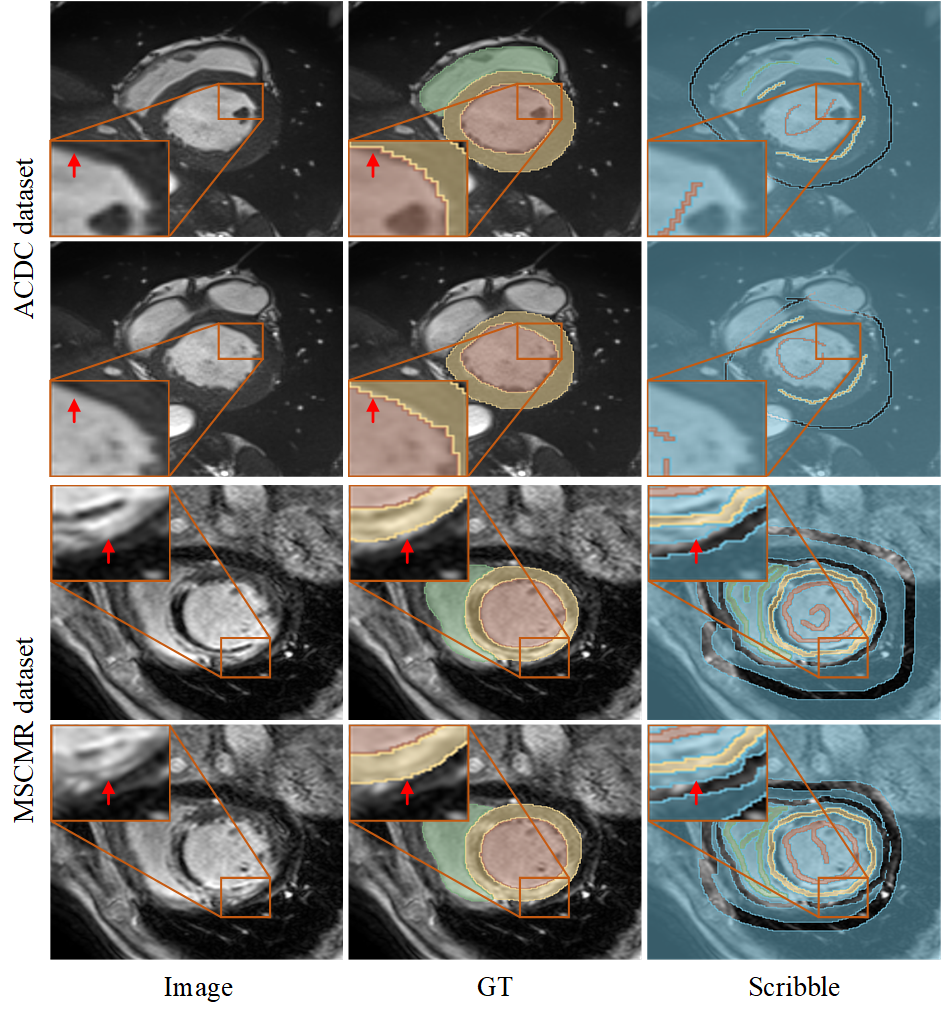}		
	\caption{Visualization of cases that appear to contain label noise near the boundaries of dense annotations.}
	\label{figure_label_noise}
\end{figure}

In medical image segmentation, the accuracy of ground truth annotations has a direct impact on model performance. Although ground truth is typically treated as the gold standard during training, it may contains noise in practice. This can be attributed to several factors: 1) subjective differences in interpretation among annotators, 2) intra-annotator inconsistencies caused by fatigue or image quality issues, and 3) the inherent complexity and blurry boundaries of certain anatomical structures, which make precise pixel-level labeling particularly challenging. Fig. \ref{figure_label_noise} presents examples from the ACDC and MSCMRseg datasets that, based on visual inspection, appear to exhibit potential noise. 

\begin{table}[t]
	\centering
	\caption{Comparisons on fully-supervised and scribble-supervised segmentation. }
	\resizebox{0.45\textwidth}{!} {
		\begin{tabular}{lccccc}
			\hline
			\multicolumn{6}{c}{ACDC dataset}                                                                                                  \\ \hline
			\multicolumn{1}{l|}{Method}     & \multicolumn{1}{c|}{Annotation} & LV            & MYO           & RV            & Avg           \\ \hline
			\multicolumn{1}{l|}{Baseline}   & \multicolumn{1}{c|}{Dense}      & 92.7$_{\pm7.7}$          & 89.7$_{\pm3.9}$          & 88.6$_{\pm8.6}$          & 90.3$_{\pm6.7}$          \\
			\multicolumn{1}{l|}{ScribbleVS} & \multicolumn{1}{c|}{Scribble}   & {92.9$_{\pm5.3}$} & ${89.4}_{\pm 2.7}$ & ${89.5}_{\pm 8.7}$& ${90.6}_{\pm 5.6}$          \\
			\multicolumn{1}{l|}{ScribbleVS} & \multicolumn{1}{c|}{Dense}      & \textbf{94.2}$_{\pm4.1}$ & \textbf{90.1}$_{\pm2.3}$ & \textbf{90.2}$_{\pm8.5}$ & \textbf{91.8}$_{\pm5.0}$ \\ \hline
			\multicolumn{6}{c}{MSCMRseg dataset}                                                                                              \\ \hline
			\multicolumn{1}{l|}{Method}     & \multicolumn{1}{c|}{Annotation} & LV            & MYO           & RV            & Avg           \\ \hline
			\multicolumn{1}{l|}{Baseline}   & \multicolumn{1}{c|}{Dense}      & 92.3$_{\pm4.2}$          & 83.3$_{\pm5.6}$          & 85.1$_{\pm6.5}$          & 86.9$_{\pm5.4}$          \\
			\multicolumn{1}{l|}{ScribbleVS} & \multicolumn{1}{c|}{Scribble}   & $\textbf{93.6}_{\pm 2.5}$ & $\textbf{85.8}_{\pm 3.4}$ & $\textbf{86.6}_{\pm 7.7}$ & $\textbf{88.7}_{\pm 4.6}$ \\
			\multicolumn{1}{l|}{ScribbleVS} & \multicolumn{1}{c|}{Dense}      & 93.3$_{\pm2.5}$          & 85.7$_{\pm3.0}$          & 85.3$_{\pm8.5}$          & 88.1$_{\pm4.7}$          \\ \hline
	\end{tabular} }
	\label{table_full_sup}
\end{table}

In contrast, scribble-based annotation significantly improves labeling efficiency and tends to focus on the core regions of target structures, making it more reliable. While it lacks detailed boundary information, this approach inherently reduces the risk of boundary-related labeling errors, which can otherwise degrade segmentation performance. 

Table \ref{table_full_sup} presents the Dice scores for fully supervised segmentation on the ACDC and MSCMRseg datasets. The results show that the ScribbleVS framework achieves superior performance using both dense and scribble-based annotations. When using dense annotations, ScribbleVS improved the average Dice score on ACDC from 90.3$\%$ to 91.8$\%$, and on MSCMRseg from 86.9$\%$ to 88.1$\%$. These findings suggest that ScribbleVS is not only well-suited for scribble supervision but also serves as a robust complement to conventional dense labeling strategies. Remarkably, with only scribble annotations, the framework achieved 90.6$\%$ and 88.7$\%$ on the two datasets, slightly outperforming the dense annotations. We hypothesize that this may be due to the presence of noise in the dense annotations of these two datasets, which can cause the model to overfit and thereby reduce its robustness.

In summary, scribble annotations strike an effective balance between labeling efficiency and segmentation accuracy, highlighting their potential for broad application in medical image segmentation. In future work, we aim to further refine scribble-based algorithms and develop more efficient, intelligent, and user-friendly annotation tools to accelerate the adoption of this promising approach.

\section{Conclusion}
\color{black}
In this work, we introduce ScribbleVS, a novel framework for medical image segmentation based on scribble-level supervision. ScribbleVS integrates two key components: the $RPD$ module, which expands the scope of supervision by propagating sparse annotations, and the $DCS$ module, which refines pseudo labels to improve adaptability. Extensive experiments on ACDC, MSCMRseg, WORD and BraTS2020 dataset demonstrate that ScribbleVS consistently outperforms existing methods, achieving accurate pixel-level segmentation using only scribble annotations. These findings underscore the potential of scribble annotations as an efficient and cost-effective alternative to dense manual labeling for medical image segmentation.

\bibliographystyle{IEEEtran}
\bibliography{reference}

\end{document}